\documentclass[11pt]{article}
\usepackage{nodalida2025}
\usepackage{times}
\usepackage[hyphenbreaks]{breakurl}
\usepackage{latexsym}
\usepackage{multirow}
\usepackage{graphicx}
\usepackage{microtype}
\usepackage[hidelinks]{hyperref}
\usepackage{booktabs}
\usepackage{adjustbox}
\usepackage[T1]{fontenc}
\hypersetup{
           breaklinks=true,
        }

\aclfinalcopy 

\title{How to Tune a Multilingual Encoder Model for Germanic Languages: \\ A Study of PEFT, Full Fine-Tuning, and Language Adapters}

\author{Romina Oji \and Jenny Kunz \\
  Dept.\ of Computer and Information Science \\
  Linköping University \\
  {\tt romina.oji@liu.se} \and {\tt jenny.kunz@liu.se}}

\date{}

\begin{document}
\maketitle
\begin{abstract}
This paper investigates the optimal use of the multilingual encoder model mDeBERTa for tasks in three Germanic languages -- German, Swedish, and Icelandic -- representing varying levels of presence and likely data quality in mDeBERTas pre-training data. We compare full fine-tuning with the parameter-efficient fine-tuning (PEFT) methods LoRA and Pfeiffer bottleneck adapters, finding that PEFT is more effective for the higher-resource language, German. However, results for Swedish and Icelandic are less consistent. We also observe differences between tasks: While PEFT tends to work better for question answering, full fine-tuning is preferable for named entity recognition. 
Inspired by previous research on modular approaches that combine task and language adapters, we evaluate the impact of adding PEFT modules trained on unstructured text, finding that this approach is not beneficial. 
\end{abstract}

\section{Introduction}

Massively multilingual encoder models like mBERT \citep{devlin-etal-2019-bert}, XLM-R \citep{conneau-etal-2020-unsupervised} and mDeBERTa \citep{he2021debertadecodingenhancedbertdisentangled} are a workhorse for NLP in many lower-resource languages. However, due to interference between languages \citep{conneau-etal-2020-unsupervised, chang2023multilingualitycurselanguagemodeling}, these models can fall short of reaching their full potential for individual target languages: Monolingual models \citep{virtanen2019multilingualenoughbertfinnish, snaebjarnarson-etal-2022-warm} and models with dedicated language modules \citep{pfeiffer-etal-2022-lifting, blevins2024breakingcursemultilingualitycrosslingual} frequently outperform them, raising the question for the best setups for different languages. 

Parameter-efficient fine-tuning (PEFT) methods, such as bottleneck adapters \citep{pmlr-v97-houlsby19a}, LoRA \citep{hu2022lora}, and prefix tuning \citep{li-liang-2021-prefix}, have emerged as an alternative to full fine-tuning of pre-trained language models. These methods preserve the model's representations and can lead to better generalisation \citep{he-etal-2021-effectiveness}. This is especially relevant for multilingual models, trained on diverse data, of which the target language only constitutes a small fraction. Fully fine-tuning them on task-specific data risks overwriting some of the multilingual capabilities. 

Language adapters -- PEFT modules trained on unstructured text independently from task fine-tuning -- have shown promise in cross-lingual transfer \citep{pfeiffer-etal-2020-mad, vidoni2020orthogonallanguagetaskadapters}. 
We explore whether language adaptation modules are beneficial even in scenarios where cross-lingual transfer is \textit{not} required, i.e., where we have in-language fine-tuning data. In addition, we use not only bottleneck (Pfeiffer) adapters but also LoRA \citep{hu2022lora}, a method that has become popular for LLMs as its parameters can be merged with the model parameters, adding no inference overhead. 

In this paper, we investigate strategies for adapting a multilingual encoder model to task data in three languages: German, Swedish, and Icelandic. 
For this, we use multilingual DeBERTa \citep{he2021debertadecodingenhancedbertdisentangled}, which is currently the best-performing model according to the ScandEval \citep{nielsen2024encodervsdecodercomparative} leaderboard for Icelandic,\footnote{https://scandeval.com/icelandic-nlu/, as of 21/10/2024.} the lowest-resourced and thus the most challenging of the three languages. 

Our findings indicate that the effectiveness of full fine-tuning versus PEFT varies by language. For German, a PEFT method consistently delivers the best results, although sometimes with marginal gains. For Swedish and Icelandic, the performance is task-dependent: PEFT is more beneficial for extractive question-answering (QA), while full fine-tuning works better for named entity recognition (NER). We hypothesise that in languages with quantitatively more limited or lower-quality representation in the pre-training data, there is less value in preserving the pre-existing representations and more value in increasing the learning capacity. In contrast, for higher-resource languages, capabilities from the pre-training phase are more impactful. Similarly, for extractive QA, pre-existing skills weigh higher, while the highly specific nature of NER benefits from full fine-tuning.

Language adapters do not provide consistent improvements in any of the tasks or languages tested. As the adaptation data we use has likely been used for pre-training the multilingual DeBERTa model, we conclude that the utilisation of this data at pre-training time has already been effective enough. Further adaptation, or specialisation, with this same data does not have a clear benefit. 

\section{Related Work}

PEFT methods not only reduce the number of trainable parameters and, consequently, memory usage in comparison to full fine-tuning, but there is also evidence suggesting that they provide better regularisation and help preserve pre-existing model capabilities. For example, \citet{he-etal-2021-effectiveness} demonstrate that adapter-based fine-tuning outperforms full fine-tuning in cross-lingual transfer setups, likely by avoiding overfitting on the source language. Similarly, prefix tuning, another PEFT method, has been shown to surpass full fine-tuning in extrapolation scenarios \citep{li-liang-2021-prefix}. 

Other works have shown the effectiveness of bottleneck-style adapters in cross-lingual transfer as post-hoc trained language modules in encoder models. \citet{pfeiffer-etal-2020-mad} show that bottleneck language adapters in the Pfeiffer architecture improve performance in NER, commonsense classification, and extractive QA. Even \citet{vidoni2020orthogonallanguagetaskadapters} report that language adapters are effective. Other research indicates that language adapters can aid in transferring knowledge to dialectal variants \citep{vamvas-etal-2024-modular} and that sharing adapters across related languages can be beneficial \citep{faisal-anastasopoulos-2022-phylogeny, chronopoulou-etal-2023-language}. However, the success of language adapters may be task-specific and difficult to measure accurately when using machine-translated evaluation data \citep{kunz-holmstrom-2024-impact}. And notably, none of the works used multilingual DeBERTa models, which may explain divergences in results. 

\section{Experimental Setup}
\subsection{Model}
We use the multilingual DeBERTa v3 model\footnote{loaded from \url{https://huggingface.co/microsoft/mdeberta-v3-base}} as the base for our experiments. 
This model contains about 86 million parameters in its backbone, and the embedding layer, with a vocabulary of 250,000 tokens, adds another 190 million parameters, bringing the total to around 278 million parameters\citep{he2021debertav3}. 
It was trained on 2.5 TB of the CC100 multilingual dataset \citep{wenzek-etal-2020-ccnet, conneau-etal-2020-unsupervised}, which includes 100 languages, including Icelandic, Swedish, and German. 

\subsection{Tasks}
\label{sec:tasks}
We evaluate the fine-tuning and language adaptation methods on three tasks: extractive question answering (QA), named entity recognition (NER), and linguistic acceptability classification. This selection is inspired by coverage in the ScandEval benchmark \citep{nielsen2024encodervsdecodercomparative} for all three languages while having structurally different tasks. 

\paragraph{QA:} For Icelandic, we use the \textit{Natural Questions in Icelandic (NQiI)} dataset, which features questions from Icelandic texts written by Icelandic speakers.\citep{snaebjarnarson-einarsson-2022-natural}. 
For Swedish, we use the Swedish portion of ScandiQA, 
which was manually translated from English \citep{nielsen2023scandeval}. 
For German, we use the human-labeled GermanQuAD dataset, which is natively German.  \citep{moller2021germanquad}. 

\paragraph{NER:} For Icelandic, we use the MIM-GOLD-NER dataset \citep{20.500.12537/42}, for Swedish, we use the Stockholm-Umeå Corpus \citep{kurtz2022sucx} and for German, we use GermanEval 2014 \citep{benikova-etal-2014-nosta}. 

\paragraph{Linguistic Acceptability: }For all three languages, we use the respective portion of ScaLA \citep{nielsen2023scandeval}, a binary classification dataset that judges the linguistic acceptability of sentences. Sentences are tagged as either grammatically correct or incorrect. This dataset is synthetically created by introducing corruptions based on the dependency trees of the sentences. 

\subsection{PEFT Methods}

We use two different PEFT methods. 
\textbf{Pfeiffer adapters} \citep{pfeiffer-etal-2021-adapterfusion, pfeiffer-etal-2020-mad} are a variation of bottleneck adapters \citep{pmlr-v97-houlsby19a}, that is, small feed-forward layers that reduce the dimensionality of the input, process it, and then expand it back to the original size. They are inserted between the layers of the transformer model, and are the only parameters that are trained. 
\textbf{LoRA} \citep{hu2022lora} approximates the original weight updates as a low-rank decomposition by learning two low-rank matrices. Instead of updating the full set of model parameters, LoRA inserts trainable low-rank matrices into the self-attention of each layer of the model and updates only those. 

\subsection{PEFT Training}

In the first step, we fine-tune individual \textit{language adapters} for Icelandic, Swedish, and German, using the masked language modeling objective. We use 250,000 samples from the CC100 dataset and train a LoRA and a Pfeiffer language adapter for each language. Our language adapters are available at \url{https://huggingface.co/rominaoji}. 

\textit{Task adapters} are fine-tuned on target-language task data with the datasets described in Section \ref{sec:tasks}. 

For all adapters, we set the LoRA rank to 8 and the $\alpha$ to 16, while for the Pfeiffer method, the reduction factor is set to 16. For the implementation, we use the \textit{adapters} library \citep{poth-etal-2023-adapters}. 

\subsection{Setups}

To find the optimal method to use mDeBERTa for the three languages, we fine-tune it using three setups: (1) \textbf{Full fine-tuning}, (2) tuning using only \textbf{task adapters}, and (3) using a \textbf{combination of language and task adapters} as in the MAD-X framework. In each setup, models are fine-tuned over five epochs. 

As PEFT models require higher learning rates than full fine-tuning due to their lower number of trainable parameters, we determine a suitable rate for each setup by testing learning rates from 1-e4 to 9e-4 for PEFT and from 1e-5 to 9e-5 for full fine-tuning. This resulted in a learning rate of 3e-4 for both the language and task adaptation methods and 2e-5 for full fine-tuning. All experiments use a linear scheduler paired with the AdamW optimiser\citep{loshchilov2017decoupled}. The code is available at \url{https://github.com/rominaoji/german-language-adapter}.

\subsection{Evaluation}

For the sake of simplicity, we only present F1 scores as the evaluation metric for all three tasks in this paper. While we have collected results on more metrics, we did not observe differences in the trends. The results are the mean of a five-fold cross-validation, with standard deviation. 

\section{Results and Discussion}

\begin{table*}[]
\centering
\adjustbox{max width=\textwidth}{%
    \begin{tabular}{lcccccccccl}\toprule
    & & \multicolumn{3}{c}{QA} & \multicolumn{3}{c}{NER} & \multicolumn{3}{c}{ScaLA}
    \\\cmidrule(lr){3-5}\cmidrule(lr){6-8}\cmidrule(lr){9-11}
    TA & LA & Icelandic & Swedish & German & Icelandic & Swedish & German & Icelandic & Swedish & German \\\midrule
    Full FT & - &   
    57.52 ± 1.50 &
    \textbf{35.08 ± 0.77} &
    73.56 ± 0.78 &
    \textit{\textbf{\textcolor{blue}{92.35 ± 0.31}}} &
    \textit{\textbf{\textcolor{blue}{87.47 ± 0.41}}} &
    84.83 ± 0.33 &
    \textbf{76.17 ± 1.32} &
    84.23 ± 1.07 &
    83.90 ± 0.82 
    
    \\\midrule
    Pfeiffer & - &
    \textbf{59.31 ± 1.14} &
    \textit{\textbf{\textcolor{blue}{35.15 ± 1.00}}} &
    75.84 ± 0.92 &
    91.37 ± 0.23 &
    86.64 ± 0.51 &
    \textit{\textbf{\textcolor{blue}{85.14 ± 0.22}}} &
    \textit{\textbf{\textcolor{blue}{76.35 ± 0.56}}} &
    \textbf{84.94 ± 1.04} &
    \textit{\textbf{\textcolor{blue}{84.53 ± 0.64}} }
    \\
    LoRA & - &
    57.65 ± 2.11 &
    34.76 ± 0.82 &
    \textit{\textbf{\textcolor{blue}{77.17 ± 0.74}}} &
    89.69 ± 0.49 &
    85.16 ± 0.32 &
    84.12 ± 0.31 &
    70.64 ± 2.78 &
    82.32 ± 1.80 &
    78.75 ± 2.34
    \\\midrule
    Pfeiffer & Pfeiffer &
    \textit{\textbf{\textcolor{blue}{60.02 ± 1.46}}} &
    \textbf{35.07 ± 0.78} &
    76.66 ± 0.55 &
    91.41 ± 0.23 &
    \textbf{86.72 ± 0.28} &
    84.77 ± 0.38 &
    75.38 ± 1.21 &
    84.68 ± 0.76 &
    \textbf{84.02 ± 0.64}
    \\
    LoRA & Pfeiffer &
    57.44 ± 1.61 &
    34.77 ± 0.60 &
    \textbf{77.13 ± 0.28} &
    89.95 ± 0.50 &
    85.11 ± 0.30 &
    84.05 ± 0.35 &
    71.31 ± 2.30 &
    82.58 ± 2.01 &
    78.85 ± 2.28
    \\
    Pfeiffer & LoRA &
    59.24 ± 0.60 &
    34.97 ± 0.82 &
    76.31 ± 0.63 &
    \textbf{91.49 ± 0.29} &
    86.50 ± 0.60 &
    \textbf{85.08 ± 0.22} &
    75.06 ± 1.41 &
    \textit{\textbf{\textcolor{blue}{84.98 ± 1.23}}} &
    83.86 ± 0.30
    \\
    LoRA & LoRA &
    57.05 ± 1.74 &
    34.40 ± 0.40 &
    77.02 ± 0.35 &
    89.64 ± 0.43 &
    85.11 ± 0.30 &
    84.08 ± 0.20 &
    71.01 ± 3.00 &
    82.97 ± 1.78 &
    78.94 ± 2.37
    \\\bottomrule
    \end{tabular}}
    \caption{Mean F1 scores over five runs with standard deviation for all tasks and languages. The first column specifies the task adaptation method (TA), and the second one the language adaptation method (LA). The respectively highest score is highlighted in bold blue italics, the runner-up in bold black. }
    \label{tab:f1}
\end{table*}

All results are presented in Table \ref{tab:f1}. We discuss the effects of different task fine-tuning strategies on different languages and tasks (§\ref{sec:ft_peft}) and finally the effect of language adapters (§\ref{sec:la}). 

\subsection{Full Fine-Tuning Versus PEFT}
\label{sec:ft_peft}

\paragraph{Tasks:} For the extractive QA tasks, we observe that PEFT methods generally outperform full fine-tuning. In German, there is a notable gap between full fine-tuning and both PEFT methods, with LoRA yielding the best results. For Icelandic, Pfeiffer adapters outperform both full fine-tuning and LoRA. For Swedish, the differences between setups are minimal. We hypothesise that for this task, the model benefits from the pre-trained representations and does not require the highest possible learning capacity to identify relevant text spans in these tasks.

In contrast, full fine-tuning is the best approach for NER tasks, outperforming the highest-performing PEFT method in Icelandic and Swedish, and performing on par with Pfeiffer adapters in German. This suggests that for this word-level task, a larger learning capacity is more crucial than preserving fine-grained capabilities from pre-training.

For ScaLA, the results are mixed. Full fine-tuning yields slightly higher scores for Icelandic, while Pfeiffer adapters perform marginally better for Swedish and German. Interpreting the performance on this task is challenging, as the dataset contains some corrupted instances that may be detectable with simple pattern-matching, while others require more fine-grained linguistic knowledge.

\paragraph{Languages:} For German, Pfeiffer adapters consistently outperform full fine-tuning in QA and ScaLA tasks, and are either on par or slightly better for NER. LoRA performs best for QA but yields lower scores in the other two tasks. This suggests that German benefits from keeping the base model intact, likely due to its relatively large representation in the pre-training dataset.

For Swedish, the performance of full fine-tuning and Pfeiffer adapters is similar across all three tasks, showing little variation.

For Icelandic, Pfeiffer adapters achieve higher scores in QA, while full fine-tuning performs better for NER. For ScaLA, both approaches produce comparable results. Icelandic's low representation in the CC-100 dataset used to train mDeBERTa might explain why it benefits less from the model’s pre-training than German. While Swedish even has a slightly larger quantitative representation than German in open CC100 dumps,\footnote{See e.g.\ \url{https://huggingface.co/datasets/statmt/cc100} as of 23/10/2024.} it is unclear if the quality of the Swedish data matches that of the German data. For lesser-resourced languages, the quality of common-crawl corpora is often lower \citep{kreutzer-etal-2022-quality, artetxe-etal-2022-corpus}, which may diminish the usefulness of pre-training for Swedish compared to German. Swedish has 13M speakers (10M L1),\footnote{\url{https://en.wikipedia.org/wiki/Swedish_language} as of 23/10/2024.} whereas German has 175M speakers (95M L1),\footnote{\url{https://en.wikipedia.org/wiki/German_language} as of 23/10/2024.} which probably makes German higher-resource than Swedish, and may lead to higher-quality representation of German. 

\paragraph{PEFT Methods:} Except for German QA, Pfeiffer adapters outperform LoRA across all tasks. This may be due to architectural differences, though it is worth noting that Pfeiffer adapters in our setup have a higher learning capacity, with 896K trainable parameters compared to LoRA’s 296K. Additionally, LoRA may require more extensive hyperparameter tuning than Pfeiffer adapters, as previous studies have shown its behavior to be unstable under certain conditions \citep{liu2024doraweightdecomposedlowrankadaptation}. A deeper exploration of how to improve LoRA’s adaptation is left for future work.

\subsection{Language Adaptation}
\label{sec:la}

Language adapters do not provide any significant benefits. When using Pfeiffer task adapters, performance remains similar whether language adapters are included or not. The only exception is Icelandic QA, where the combination of a Pfeiffer language adapter and a Pfeiffer task adapter achieves a slightly higher score compared to the best setup without language adapters. However, the difference is small and possibly due to result variability, as it falls within a standard deviation.

With LoRA task adapters, language adaptation methods sometimes result in a noticeable performance drop, suggesting potential interference. While prior work, such as \citet{pfeiffer-etal-2020-mad}, reported improvements in similar tasks, their study focused on cross-lingual transfer, where no task data from the target language was available. In contrast, our setups use task data from the target language, and all the languages are present in the model's pre-training data. In addition, we use mDeBERTa-v3, which reportedly performs better for the languages in question than the XLM-R \citep{conneau-etal-2020-unsupervised} and multilingual BERT \citep{devlin-etal-2019-bert} models that most other papers including \citet{pfeiffer-etal-2020-mad} use. 
These factors likely contribute to the fact that language adapters are unnecessary in our setup.

\section{Conclusion}

We compared the performance of the multilingual encoder model mDeBERTa across three task adaptation setups: full fine-tuning, bottleneck (Pfeiffer) adapters, and LoRA. Based on our evaluations across three tasks and three languages, we found that the choice of the best method is both task- and language-dependent. Specifically, extractive QA tasks benefit from PEFT methods, while NER gets better results with full fine-tuning. For German, a higher-resourced language, PEFT consistently achieves higher scores. This suggests that the model benefits from fine-grained information learned during pre-training if coverage and (or) quality of the language data in the pre-training corpus are sufficiently high. In contrast, for lower-resourced languages, the increased learning capacity of full fine-tuning proves more advantageous.

We also tested language adaptation with Pfeiffer adapters and LoRA on unstructured text data before task adaptation. However, language adapters did not show any benefit. Access to target-language task data appears to dispense with the need for them, at least in our experiments where all languages are included in the pre-training data. 

In future work, we aim to further explore the conditions under which PEFT methods versus full fine-tuning are most effective. We plan to investigate additional PEFT methods and tasks and optimise the LoRA setup, which may not have reached its full potential in our experiments.

\section*{Acknowledgments}

We thank our colleagues Kevin Glocker, Kätriin Kukk, Julian Schlenker, Marcel Bollmann and Noah-Manuel Michael for valuable discussions at all stages of this project and feedback on earlier drafts, and the anonymous reviewers for their constructive feedback and insightful suggestions. 

This research was supported by TrustLLM funded by Horizon Europe GA 101135671 and the National Graduate School of Computer Science in Sweden (CUGS). 
It was partially supported by the Wallenberg AI, Autonomous Systems and Software Program (WASP) funded by the Knut and Alice Wallenberg Foundation. 
The computations were enabled by the Berzelius resource provided by the Knut and Alice Wallenberg Foundation at the National Supercomputer Centre. 

\bibliographystyle{acl_natbib}
\bibliography{nodalida2025}

\end{document}